\documentclass[10pt,journal,compsoc]{IEEEtran}
\usepackage{graphicx}
\usepackage{epsfig}
\usepackage{epstopdf}
\usepackage{float}

\ifCLASSOPTIONcompsoc
  \usepackage[nocompress]{cite}
\else
  \usepackage{cite}
\fi
\ifCLASSINFOpdf
\else
\fi
\begin{document}
\title{Online Behavioral Analysis with Application to Emotion State Identification}

%
%

\author{Lei~Gao,~\IEEEmembership{student Member,~IEEE,}
        Lin~Qi,
        and~Ling~Guan,~\IEEEmembership{Fellow,~IEEE}
\IEEEcompsocitemizethanks{\IEEEcompsocthanksitem L. Gao and L. Guan is with the Department of Electrical and Computer Engineering, Ryerson University, Canada.\protect\\
E-mail: iegaolei@gmail.com; lguan@ee.ryerson.ca.
\IEEEcompsocthanksitem L. Qi is with the School of Information Engineering, Zhengzhou University, China.\protect\\
Email: ielqi@zzu.edu.cn}}
\IEEEtitleabstractindextext{%
\begin{abstract}
In this paper, we propose a novel discriminative model for online behavioral analysis with application to emotion state identification. The proposed model is able to extract more discriminative characteristics from behavioral data effectively and find the direction of optimal projection efficiently to satisfy requirements of online data analysis, leading to better utilization of the behavioral information to produce more accurate recognition results.
\end{abstract}

}
\maketitle

\IEEEdisplaynontitleabstractindextext

%
\IEEEpeerreviewmaketitle

\IEEEraisesectionheading{\section{Introduction}\label{sec:introduction}}

\IEEEPARstart{W}{ith} proliferation of web applications, such as search engine, e-education, e-commerce, social networking service and online gaming, more and more behavioral information are available online. Therefore, the need has arisen for a more natural communication interface between humans and web through online behavioral data analytics. To make the online behavioral data analytics more natural and friendly, it would be beneficial to give web the ability to recognize situations similar to how humans do.\\\indent
Thanks to the recent advancement in science and technology, behavioral data analytics has been advanced at a rapid speed. For example, users can use hand gestures for expression of their feelings and notifications of their thoughts, providing an attractive and natural interface to the web. Benefitting from the depth images, action analysis has been applied to autonomous video surveillance, video retrieval and human computer interaction. As another dimension of human behavior analysis, emotion state identification contributes to applications such as learning environment, entertainment, educational software and others.\\\indent Generally, online behavioral analytics is studied for two major purposes, \textit{understanding} and \textit{prediction}. In terms of understanding, machine learning algorithms such as principal component analysis (PCA), linear discriminant analysis (LDA), support vector machines (SVM) are adopted. On the other hand, Bayesian, Neutral network and hidden Markov model are normally considered to address the prediction problem. Since a predictive model (e.g. probabilistic or neural network based) does not necessarily need to be understood by a human, the focus of this paper is developing a discriminative model to support understanding of human behaviors and mobility patterns.\\\indent
Although researches in online behavioral analysis have advanced rapidly in recent years, realistically emulating the behavioral analysis capacity of the human brain is still far from mature. The main reason is that human brain is a natural behavioral analysis system, performing the task by studying multi-modal behavioral information from the different sensory modalities, such as sight, sound, touch, smell, self-motion and taste to have meaningful perceptual experiences. During the last few years, behavioral analysis was improved (reaction time and accuracy) when the objects were presented with multi-modal features compared to single modal features alone [1], suggesting that studying multi-modal information is a promising direction to explore in behavioral analysis. When one modality fails or is not good enough to determine a certain behavior, the other modalities may help to improve the performance.\\\indent Although the study of  multi-modal data for online behavioral analysis has been drawing attentions of the research community [2-3], it faces major challenges in the identification of the inherent relationship between different modalities, and the design of a fusion strategy that can effectively utilize the discriminatory information presented in different channels. \\\indent In this paper, we present a discriminative model for online behavioral analytics with application to emotion state identification. At first, it finds projected directions to maximize the correlation among multiple behavioral data in order to identify the inherent relationship between different modalities. Second, based on the proposed model, we verify that the best performance by discriminative representation achieves when only a small fraction of the data needs to be analyzed in numerous popular applications such as emotion recognition, digit and English character recognition. The effectiveness of the proposed model is demonstrated using comparison with serial fusion [4] and methods based on similar principles such as CCA [5], DCCA [6] and MCCA [7].
\section{The Discriminative Model}
In this section, we introduce a discriminative model to identify the inherent relationship and extract discriminatory representations between different modalities. The advantages of the proposed model for multi-modal behavioral data fusion rest on the following facts: 1) the correlation among the variables in multiple channels is taken as the metric of the similarity between the variables; 2) the within-class similarity and the between-class dissimilarity are considered jointly to extract discriminatory information.\\\indent Given \emph{$ P $} sets of zero-mean random behavioral features ${x_1} \in {R^{{m_1} \times n}},{x_2} \in {R^{{m_2} \times n}}, \cdots {x_P} \in {R^{{m_P} \times n}}$ for \textit{c} classes and $Q = {m_1} + {m_2} +  \cdots {m_P}$. Concretely, the discriminative model aims to seek the projected vectors $\omega  = {[{\omega _1}^T,{\omega _2}^T \cdots {\omega _P}^T]^T}$ $({\omega_1} \in {R^{{m_1} \times Q}},{\omega_2} \in {R^{{m_2} \times Q}}, \cdots {\omega_P} \in {R^{{m_P} \times Q}})$ for fused features extraction so that the within-class correlation is maximized and the between-class correlation is minimized. Specifically, it is formulated as the following optimization problem:
\begin{equation} \mathop {\arg \max \rho }\limits_{{\omega _1},{\omega _2} \cdots {\omega _P}}  = \frac{1}{{P(P - 1)}}\sum\limits_{\scriptstyle k,m = 1\hfill\atop \scriptstyle{\rm{ }}k \ne m\hfill}^P {{\omega _k}^T\mathop {{C_{{x_k}{x_m}}}}\limits^ \sim  {\omega _m}} {\rm{  (}}k \ne m{\rm{)}} \end{equation}
Subject to
\begin{equation} \sum\limits_{k = 1}^P {{\omega _k}^T{C_{{x_k}{x_k}}}{\omega _k}}  = P \end{equation}
where $\mathop {{C_{{x_k}{x_m}}}} \limits^ \sim  = {C_w} -  {C_b},{\rm{ }}{C_{{x_k}{x_k}}} = {x_k}{x_k}^T$. ${C_w}$ and ${C_b}$ denote the within-class and between-class correlation matrixes of \emph{$ P $} sets, respectively. The definition of $\mathop {{C_{{x_k}{x_m}}}}\limits^ \sim$ indicates that the optimal solution to (1) achieves by simultaneously minimizing the between-class correlation and maximizing the within-class correlation.\\\indent Based on the mathematical analysis in Appendix A, $C_w$ and $C_b$ can be explicitly expressed in equation (3) and (4).
\begin{equation}
{{C_w} = {x_k}A{x_m}^T} \end{equation}
\begin{equation} {C_b} =  - {x_k}A{x_m}^T \end{equation}
\begin{equation} \ A = \left[ {\left( {\begin{array}{*{20}{c}}{{H_{{n_{i1}} \times {n_{i1}}}}}& \ldots &0\\
 \vdots &{{H_{{n_{il}} \times {n_{il}}}}}& \vdots \\
0& \ldots &{{H_{_{{n_{ic}} \times {n_{ic}}}}}}
\end{array}} \right)} \right] \in {R^{n \times n}} \end{equation}
where ${n_{il}}$ is the number of samples in the \textit{l}th class of the set $x_i$ and ${H_{{n_{i{1}}} \times {n_{i1}}}}$ is in the form of ${n_{i1}} \times {n_{i1}}$ with unit values for all the elements. \\ Substituting equation (3) and (4) into (1) yields:
\begin{equation} \frac{{1}}{{P - 1}}(C - D)\omega  = \rho D\omega \end{equation}
where
\begin{equation} \ C = \left[ {\left( {\begin{array}{*{20}{c}}
{{x_1}{x_1}^T}& \ldots &{{x_1}A{x_P}^T}\\
 \vdots & \ddots & \vdots \\
{{x_P}A{x_1}^T}& \cdots &{{x_P}{x_P}^T}
\end{array}} \right)} \right] \end{equation}

\begin{equation} \ D = \left[ {\left( {\begin{array}{*{20}{c}}
{{x_1}{x_1}^T}& \ldots &0\\
 \vdots & \ddots & \vdots \\
0& \cdots &{{x_P}{x_P}^T}
\end{array}} \right)} \right] \end{equation}
\begin{equation}
\rho  = \left( {\begin{array}{*{20}{c}}
{{\rho _1}}& \ldots &0\\
 \vdots & \ddots & \vdots \\
0& \cdots &{{\rho _Q}}
\end{array}} \right)
 \end{equation}
\begin{equation} \omega  = [{\omega ^T}_1,{\omega ^T}_2, \cdots {\omega ^T}_P]^T \end{equation}
Based on the definition of $\mathop {{C_{{x_k}{x_m}}}}\limits^ \sim  $ in equation (1) , the value of ${\rho _i}$ $(i = 1,2,...,Q)$ in equation (9) plays a critical role in evaluating the relationship between within-class and between-class correlation matrixes. When the value of ${\rho _i}$ is greater than zero, the corresponding projected vector contributes positively to the discriminative power in classification while the projected vector corresponding to the non-positive values of ${\rho _i}$ would result in reducing the discriminative power in classification. Clearly, the solution obtained is the eigenvector associated to the largest eigenvalues in equation (6). \\\indent It is known that the time taken greatly depends on the computational process of eigenvalue-decomposition. When the rank of eigen-matrix is very high, the computation of eigenvalues and eigenvectors will be time-consuming. It is a big challenge to satisfy the requirement of online behavioral analysis. From studying the properties of within-class correlation matrix $C_w$ and between-class correlation matrix $C_b$, an important characteristic of the proposed model is discovered: the number of projected dimension \emph{d} corresponding to the optimal recognition accuracy is smaller than or equal to the number of classes \textit{c}:
\begin{equation} \ d \le c \end{equation}
The derivation of (11) is given in Appendix B. Therefore, we only need to calculate the first \emph{c} projected dimensions of the discriminative model to obtain the discriminatory representations, eliminating the need of computing the complete transformation process. Specifically, if the dimension of features space in fusion equals to \emph{M}, the computational complexity of the proposed method is in the order of \emph{O}(\emph{M}*\emph{c}), instead of \emph{O}(\emph{M}*\emph{M}) as the other transformation based methods would require, to find the optimal recognition accuracy. Thus, inequality (11) is particularly significant when \emph{c} is small compared with the dimension of feature space such as emotion recognition, digit recognition, English character recognition and many others, where \emph{c} ranges from a handful to a couple of dozens, but the dimension of feature space could be of hundreds or even thousands. \\\indent  Now, we will graphically verify the effectiveness of (11) for selecting optimal projection in information fusion. In general, the solutions to a large number of multi-modal information fusion methods are obtained by utilizing the algorithm of matrix transformation. Some examples are PCA, CCA, Cross-Modal Factor Analysis (CFA), Entropy Component analysis (ECA) and their kernel versions. The solutions to matrix transformation are usually the eigenvector associated with the eigenvalue in a form similar to that of equation (6):
\begin{equation}
\frac{1}{{P - 1}}inv(D)*(C - D)\omega  = \rho \omega
\end{equation}
where \textit{inv()} refers to the inverse transform of a matrix. However, unless the covariance matrices \textit{D} have full rank, the block matrix in equation (12) is singular. An approach [14] to dealing with singular covariance matrices and to controlling complexity is to add a multiple of the identity matrix $\lambda {\rm I}(\lambda  > 0)$ to \textit{D}. Thus, the generalized form of equation (12) can be written as:
\begin{equation}
\frac{1}{{P - 1}}inv({D^ + })*(C - D)\omega  = \eta \omega
\end{equation}
where
\begin{equation}
\begin{array}{l}
 {D^ + } = \left\{ {\begin{array}{*{20}{c}}
   {D\qquad \quad             {\rm{when}}\quad D\quad {\rm{is}}\quad {\rm{a}}\quad {\rm{full}}\quad {\rm{rank}}\quad {\rm{matrix}}}  \\
   {D + \lambda {\rm{I}}\quad {\rm{when}}\quad D\quad {\rm{is}}\quad {\rm{a}}\quad {\rm{singular}}\quad {\rm{matrix}}}  \\
\end{array}} \right. \\
 \eta  = \left( {\begin{array}{*{20}{c}}
   {{\eta _1}} &  \ldots  & 0  \\
    \vdots  &  \ddots  &  \vdots   \\
   0 &  \cdots  & {{\eta _Q}}  \\
\end{array}} \right) \\
 \end{array}
\end{equation}
In equation (13), $\eta_i$  is the criterion to seek the projected vectors for feature extraction. Hence, the value of $\eta_i$ is the key parameter to the effect of selecting features. A larger $\eta_i$ corresponds to the more discriminative feature, while a smaller $\eta_i$ corresponds to the less discriminative feature. \\\indent Thus, the optimal dimension of multi-modal information fusion results can be obtained by graphically plotting $J(\eta_q)$ vs $i$ with $J(\eta_q)$ defined by
\begin{equation}
J(\eta_q) = \sum\limits_{i = 1}^q {{\eta _i}}
\end{equation}
where $ q = 1,2,...,Q$ and ${{\eta _i}}$ is the \textit{i}th eigenvalue of equation (13).
\section{Emotion State Identification}
As a key behavior of humans, emotion plays a central role in our daily social interactions and activities. It reflects an individual's state of mind in response to the internal and external stimuli. Web recognition of human emotion has become an increasingly important research topic for accomplishing the goal of building a more natural and friendly communication interface between humans and the web. \\\indent Since visual and audio information are considered two major indicators of human affective state, and thus play a leading role in emotion recognition, substantial studies have been conducted in human emotion state identification in the past few decades. Facial Action Coding System (FACS) [15], Aligned Cluster Analysis (ACA) [16] and dimensional emotion system [17] are the three major models to address this problem. Although bimodal analysis has gained momentum in recent years, the majority of the works focus on speech alone, or facial expression only. However, as shown in [10], some of the emotion recognition tasks are audio dominant, while the others are visual dominant. The complementary relationship of these two modalities should be further explored to further improve the performance. A wide investigation on the dimensions of emotions reveals that at least six emotions are universal. The set of six principal emotions proposed by Ekman [15] is: happiness(HA), sadness(SA), anger(AN), fear(FE), surprise(SU), and disgust(DI), which are the focus of study in this paper. \\\indent State-of-the-art in multi-modal emotion state identification can be found in [8] and [9] to identify the intrinsic relationship among different modalities respectively. However, selecting the discriminatory representation in the fused space for effective recognition remains a challenging open problem. In what follows, we examine the performance of the proposed model in emotion state identification from audiovisual signals.
\subsection{Audio Feature}
For emotional audio, a good reference model is the human hearing system. Currently, Prosodic, MFCC and Formant Frequency (FF) are widely used in audio emotion recognition [18-19]. As our goal is to simulate human perception of emotion, and identify quality features that can convey the underlying emotions in speech regardless of the language, identity, and context, we investigate the use of all these three types of features which are summarized as follows:\\
1)  25-dimensional Prosodic features used in [10].\\
2)  65-dimensional MFCC features: the mean, median, standard deviation, max, and range of the first 13 MFCC coefficients.\\
3)  15-dimensional Formant frequency features: the mean, median, standard deviation, max and min of the first three Formant Frequencies.\\
\subsection{Visual Feature}
Since Gabor wavelet features have been shown to be effective to represent human facial space[12], in this paper, the algorithm proposed in [13] is used to construct the Gabor filter bank which consists of
filters in 4 scales and 6 orientations. To reduce computational complexity, we calculate the mean, standard deviation and median of the magnitude of the transform coefficients of each filter as the features, including\\
1) 24-dimensional Gabor transformation feature with the mean of the transform coefficients of each filter used in [13].\\
2) 24-dimensional Gabor transformation feature with the standard deviation of the transform coefficients of each filter used in [13].\\
3) 24-dimensional Gabor transformation feature with the median of the transform coefficients of each filter used in [13].\\\indent
Note, in all experiments, the features are first transformed using the proposed model. Subsequently, the newly generated features, which represent the multi-modal information among different patterns, are concatenated into projected vectors for classification with the algorithm of nearest neighbour. In order to demonstrate the effectiveness of the proposed method, we also implemented the serial fusion, CCA, MCCA, DCCA for comparison. A general block diagram of the proposed system is illustrated in Fig. 1, in which the + with circle means fusing different features together. Note, since Ekman's six basic emotional states are used in the work, \emph{c} equals to 6 and the dimension of features (\emph{M}=177) is equal to dimension of audio features (105) plus dimension of visual features (72). The ratio of \emph{O}(\emph{M}*\emph{c}) to \emph{O}(\emph{M}*\emph{M}) is about 1:30, and application of the proposed method indeed significantly reduces the computational complexity compared with the other transformation based methods for the problem on hand. To further show the efficiency of the proposed method, we also investigate the actual running time of the proposed method and that of the MCCA. All experiments are performed on a PC with windows 7 operation system, Intel i7-3.07GHz CPU \& 10 G RAM and are always coded in MATLAB language. The running time of the proposed method is 129.43s while that of MCCA is 11043s on RML Database. The ratio of computational times is 129.43:11043 = 1: 85.3. For eNT Database, the running time of the proposed method is 224.3s while that of MCCA is 15048s. The ratio of computational time is 224.3:15048 = 1: 67 . Therefore, the actual time saving by the proposed method on the two datasets clearly demonstrated the method¡¯s efficiency again.
\begin{figure*}[t]
\centering
\includegraphics[height=3.4in,width=6.0in]{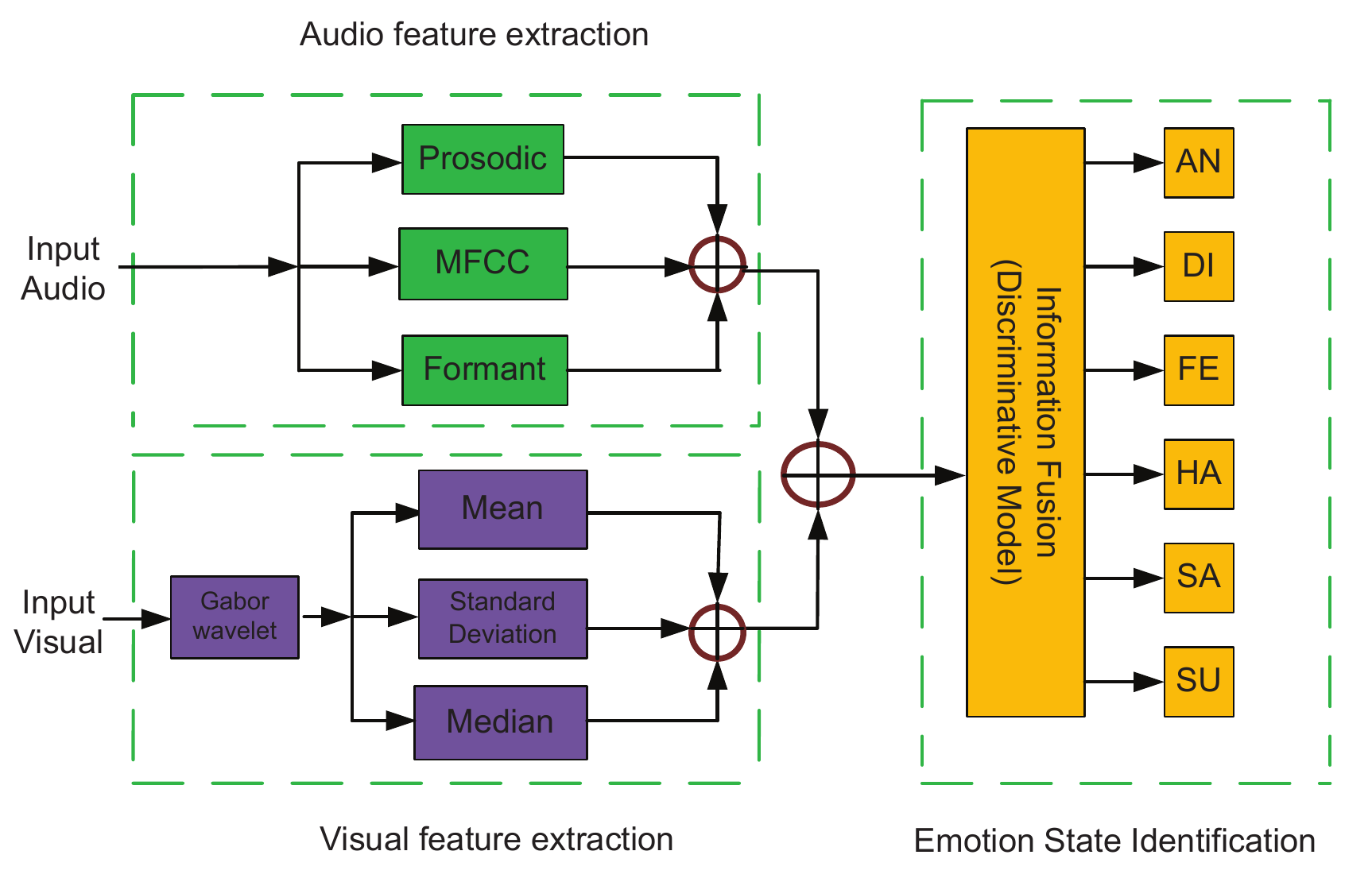}\\ Fig.1 The block diagram of the proposed system\\
\end{figure*}
\section{Experimental results and analysis}
To evaluate the effectiveness of the proposed model and criterion $J(\eta_q)$, we conduct experiments on Ryerson Multimedia Lab(RML) and eNTERFACE(eNT) audiovisual databases[11], respectively.\\\indent For audiovisual fusion based emotion state identification, visual features are extracted from the key frame image of videos, where the highest speech amplitude is found. The planar envelope approximation method in the HSV color space is used for face detection. To reduce the high memory requirement of the proposed model, 288 video clips of six basic emotions are selected for capturing the change of audio and visual information with respect to time simultaneously from RML database. Among them, 192 clips are chosen for training set and 96 for evaluation. For eNTERFACE database, 360 clips are chosen for training set and 96 for evaluation. As a benchmark, the performances of using prosodic, MFCC, formant frequency, mean, standard deviation and median features in emotion recognition are first evaluated, which are shown as TABLE 1.
\begin{table}[h]
\footnotesize
\renewcommand{\arraystretch}{1.5}
\caption{\normalsize{Results of Emotion Recognition with Single Feature}}
\setlength{\abovecaptionskip}{0pt}
\setlength{\belowcaptionskip}{10pt}
\centering
\tabcolsep 0.073in
\begin{tabular}{cc}
\hline
Single Feature & Recognition Accuracy\\
\hline
Prosodic(RML) &53.13\%\\
MFCC(RML) &47.92\%\\
Formant Frequency(RML) &29.17\%\\
Prosodic(eNT) &55.21\%\\
MFCC(eNT) &39.58\%\\
Formant Frequency(eNT) &31.25\%\\
Mean(RML) &60.42\%\\
Standard Deviation(RML) &67.71\%\\
Median(RML) &57.29\%\\
Mean(eNT) &75.00\%\\
Standard Deviation(eNT) &80.21\%\\
Median(eNT) &72.92\%\\
\hline
\end{tabular}
\end{table}\\
From TABLE 1, it shows that visual-based features achieve better recognition accuracy than audio-based. Among them, the prosodic features in audio and standard deviation of Gabor Transform coefficients in visual images could result in better performances in emotion recognition compared with other features. Therefore, in the following experiments, we will use prosodic and standard deviation for the methods of CCA and DCCA in audiovisual-based fusion. In addition, the results of serial fusion on all the six audiovisual features are investigated, and the overall recognition accuracy is 30.28\% for RML database and 35.42\% for eNTERFACE database. The performances by the methods of CCA(yellow line), MCCA(red line), DCCA(blue line), audio-based multi-feature discriminative model(magenta line), visual-based multi-feature discriminative model(cyan line) and audiovisual-based discriminative model(green line) are shown in Fig. 2. The calculation of $J(\eta_q)$ with the proposed model for audiovisual-based emotion state identification is shown as Fig.3.
\begin{figure*}[t]
\centering
\includegraphics[height=3.5in,width=6.5in]{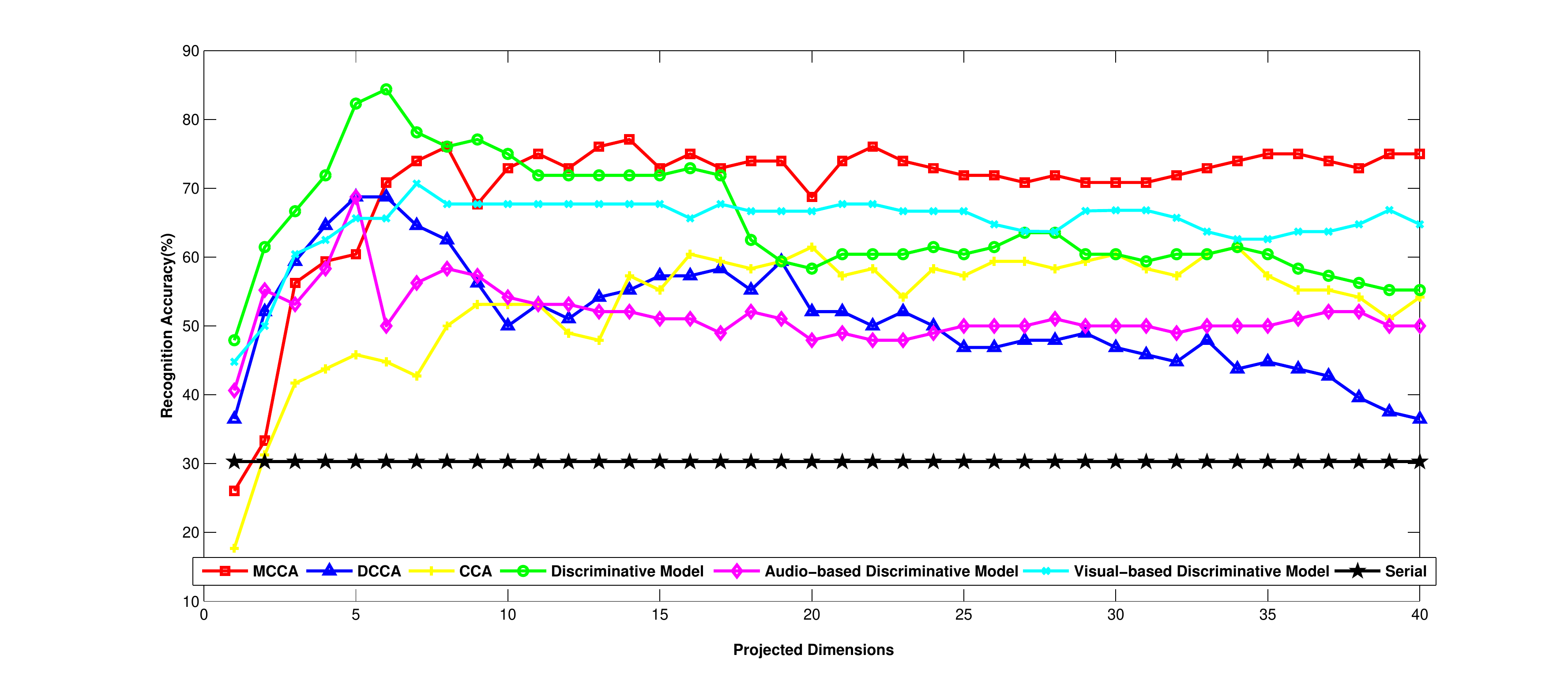}\\ Fig.2(a) Discriminative Model for Audiovisual emotion state identification experimental results with different methods on RML Database\\
\end{figure*}
\begin{figure*}[t]
\centering
\includegraphics[height=3.5in,width=6.5in]{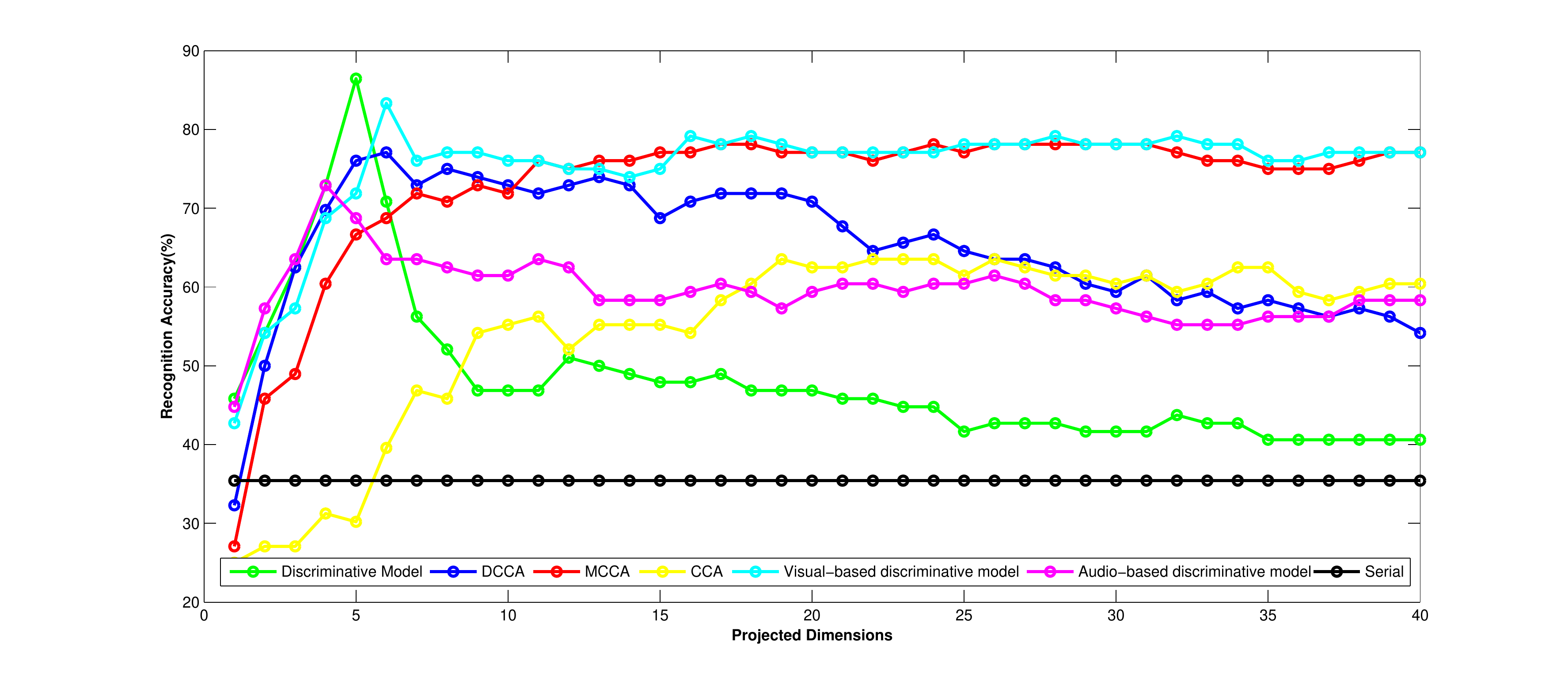}\\ Fig.2(b) Discriminative Model for Audiovisual emotion state identification experimental results with different methods on eNTERFACE Database\\
\end{figure*}

\begin{figure*}[t]
\centering
\includegraphics[height=3.5in,width=6.5in]{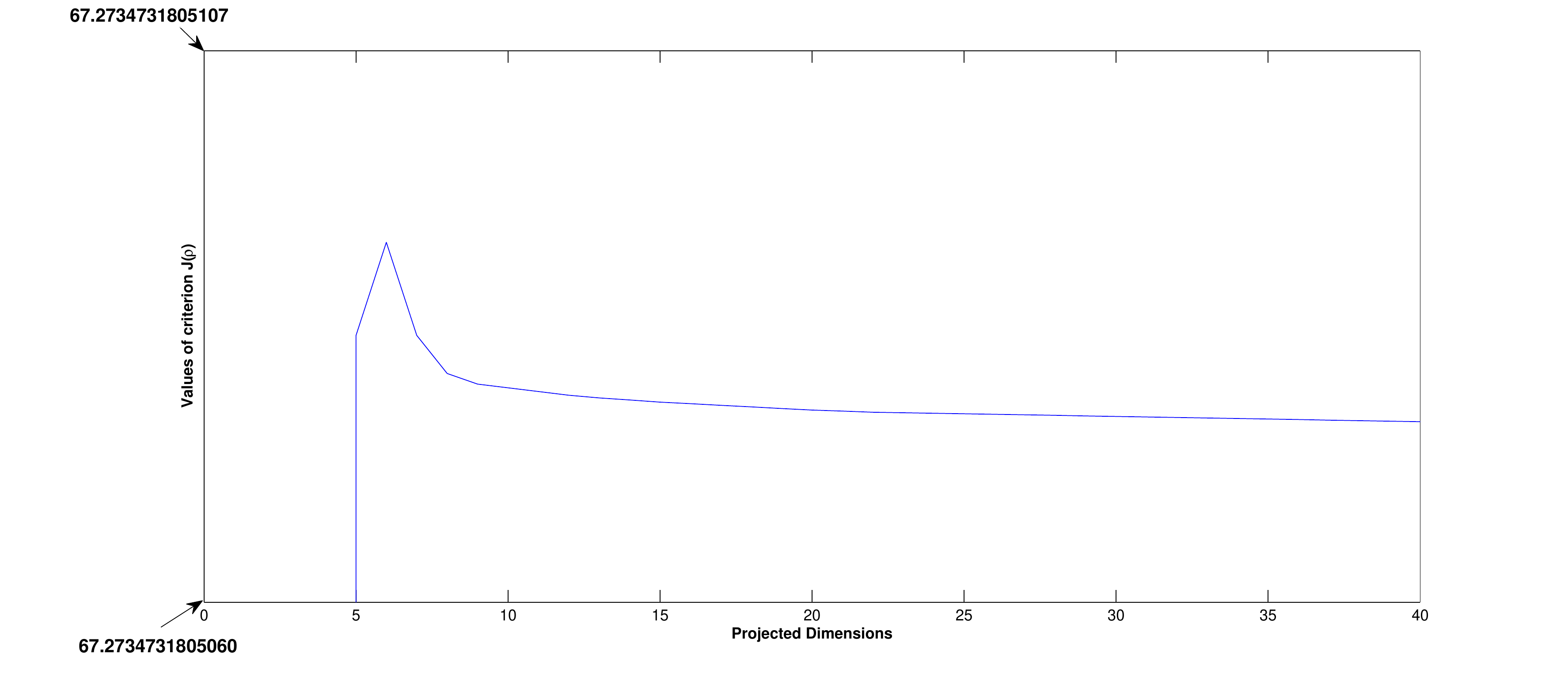}\\ Fig.3(a) The calculation of $J(\eta_q)$ with the discriminative model for audiovisual-based emotion state identification on RML Database\\
\end{figure*}
\begin{figure*}[t]
\centering
\includegraphics[height=3.5in,width=6.5in]{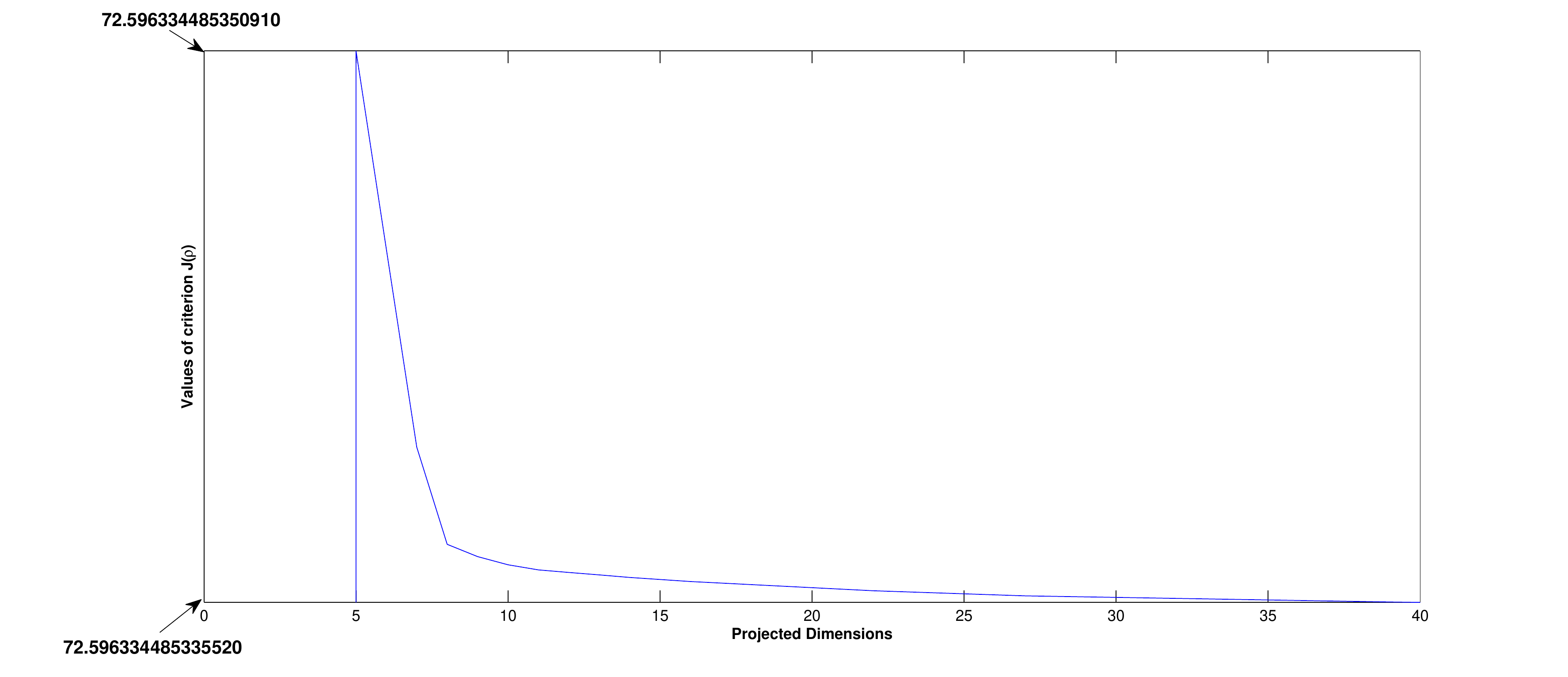}\\ Fig.3(b) The calculation of $J(\eta_q)$ with the discriminative model for audiovisual-based emotion state identification on eNTERFACE Database\\
\end{figure*}
From Fig.2, clearly, the discrimination power of the proposed model provides a more effective modeling of the relationship between multi-modal audiovisual information fusion. The fusion of multiple audio and visual information indeed enhances the performance of emotion state identification, achieving much better results than the methods compared in all cases. Since visual-based features achieves better results than audio-based features as shown in TABLE 1, visual-based discriminative model(cyan line) arrives better recognition accuracy than the methods of audio-based discriminative model(magenta line) and DCCA(blue line), which only fuses prosodic and standard deviation features. Moreover, an important finding of the researches is that, the exact location of optimal recognition performance occurs when the number of projected dimension is smaller than or equals to the number of classes \textit{c} as shown in TABLE 2. Note that, when the number of projected dimension is more than number of classes \textit{c}, the non-positive values of ${\rho _i}$ would result in reducing the discriminative power and recognition accuracy in classification at the same time. Therefore, the significance here is that, we only need to calculate the first \textit{c} projected dimensions of the discriminative model to obtain the discriminatory representations, eliminating the need of computing the complete transformation processes associated with most of the other methods. This discovery substantially reduces the computational complexity to satisfy the requirement of online processing.
\begin{table}[h]
\scriptsize
\renewcommand{\arraystretch}{1.5}
\caption{\normalsize{The relation between optimal recognition accuracy and projected dimension by different methods}}
\setlength{\abovecaptionskip}{0pt}
\setlength{\belowcaptionskip}{10pt}
\centering
\tabcolsep 0.01in
\begin{tabular}{ccc}
\hline
Method & Optimal Accuracy & Dimension(Number)\\
\hline
Discriminative Model(RML) &85.42\% & 6\\
Visual-based Discriminative Model(RML) &71.88\% & 6\\
Audio-based Discriminative Model(RML) &69.79\% & 5\\
DCCA(RML) &69.79\% & 6\\
MCCA(RML) &78.13\% & 14\\
CCA(RML) &64.58\% & 34\\
Discriminative Model(eNT) &88.54\% & 5\\
Visual-based Discriminative Model(eNT) &86.45\% & 6\\
Audio-based Discriminative Model(eNT) &75.00\% & 4\\
DCCA(eNT) &77.08\% & 6\\
MCCA(eNT) &78.13\% & 23\\
CCA(eNT) &67.17\% & 26\\
\hline
\end{tabular}
\end{table}\\
 Fig. 3 graphically illustrates the relationship between optimal projected dimensions and the recognition performance using the proposed criterion. In the figure, criterion $J(\eta_q)$ reaches the maximum when the projected dimension is 6 for RML database which is equal to the number of classes (\textit{c}=6). Similarly, the dimension of 5 is observed for the eNTERFACE database. The graphical presentation again confirms nicely with the mathematical analysis presented in Section 3.

\section{Conclusions}
In this paper, we proposed a discriminative model for online behavioral analysis and applied to emotion state identification. For the proposed model, not only the correlation from different channels is taken as the metric of the similarity between the variables, but also the within-class similarity and the between-class dissimilarity are taken into consideration for extracting the discriminatory representation. Experiments show that it outperforms serial fusion and methods based on similar principles such as CCA, MCCA and DCCA. Although we focus on the emotion state identification in this paper, the generic nature of the method enable it to be applied to other behavioral such as actions and gestures, or the combination of several behaviors. The fact that the best performance by the proposed discriminative representation can be accurately predicted offers an intuitive way of finding the optimal or near-optimal dimension of the features in the projected space in transformation based information fusion.

\appendix[ A. Proof of Equation (3) and (4)]
Let
\begin{small}
\begin{equation}
{x_i} = [{x_{i1}}^{(1)},{x_{i2}}^{(1)} \cdots {x_{i{n_1}}}^{(1)}, \cdots {x_{i1}}^{(c)},{x_{i2}}^{(c)} \cdots {x_{i{n_c}}}^{(c)}] \in {R^{{m_i} \times n}}
\end{equation}
\end{small}
\begin{equation}
{e_{{n_{il}}}} = [\underbrace {0,0, \cdots 0,}_{\sum\limits_{u = 1}^{l - 1} {{n_{iu}}} }\underbrace {1,1, \cdots 1}_{{n_{il}}}\underbrace {0,0, \cdots 0}_{n - \sum\limits_{u = 1}^l {{n_{iu}}} }]
\end{equation}
\begin{equation}
{\bf{1}} = {[1,1, \cdots 1]^T} \in {R^n}
\end{equation}
where \textit{i} is the number sequence of the random behavioral features, ${x_{ij}}^{(m)}$ denotes the \textit{j}th sample in the \textit{m}th class, and ${n_{il}}$ is the number of samples in the \textit{l}th class of the set $x_i$.
\begin{equation}
\sum\limits_{l = 1}^c {{n_{il}}}  = n
\end{equation}
where \textit{c} is the total number of classes, and \textit{n} is the total number of samples. \\ Note that, as the random features satisfy the property of zero-mean, it can be shown that:
\begin{equation}
{x_i}  {\bf{1}} = 0
\end{equation}
The within-class correlation matrix $C_w$ and between-class matrix $C_b$  can be written as:
\begin{equation} \begin{array}{l}
{C_w} = \sum\limits_{l = 1}^c {\sum\limits_{h = 1}^{{n_{kl}}} {\sum\limits_{g = 1}^{{n_{ml}}} {{x_{kh}}^{(l)}{x_{mg}}^{(l)T}} } } \\
{\rm{    }} = \sum\limits_{l = 1}^c {({x_k}{e_{{n_{kl}}}}){{({x_m}{e_{{n_{ml}}}})}^T}} \\
{\rm{    }} = {x_k}A{x_m}^T
\end{array} \end{equation}
\begin{equation} \begin{array}{l}
{C_b} = \sum\limits_{l = 1}^c {\sum\limits_{\scriptstyle q = 1\hfill\atop
\scriptstyle l \ne q\hfill}^c {\sum\limits_{h = 1}^{{n_{kl}}} {\sum\limits_{g = 1}^{{n_{mq}}} {{x_{kh}}^{(l)}{x_{mg}}^{(q)T}} } } } \\
{\rm{    }} = \sum\limits_{l = 1}^c {\sum\limits_{q = 1}^c {\sum\limits_{h = 1}^{{n_{kl}}} {\sum\limits_{g = 1}^{{n_{mq}}} {{x_{kh}}^{(l)}{x_{mg}}^{(q)T} - } } } } \sum\limits_{l = 1}^c {\sum\limits_{h = 1}^{{n_{kl}}} {\sum\limits_{g = 1}^{{n_{ml}}} {{x_{kh}}^{(l)}{x_{mg}}^{(l)T}} } } \\
{\rm{    }} = ({x_k}{\bf{1}}){({x_m}{\bf{1}})^T} - {x_k}A{x_m}^T\\
{\rm{    }} =  - {x_k}A{x_m}^T
\end{array} \end{equation}

\appendix[ B. Proof of Equation (11)]
From equation (5), the rank of matrix \emph{A} satisfies
\begin{equation} \ rank(A) \le c \end{equation}
Then, equation (23) leads to:
\begin{equation} \ rank({x_i}A{x_j}^T) \le \min ({r_i},{r_A},{r_j}) \end{equation}
where ${r_i},{\rm{  }}{r_A},{\rm{  }}{r_j}$ are the ranks of matrixes ${x_i},A,{x_j}$ (
$ i,j \in [1,2,3,...,P] $), respectively.\\
Due to the fact that $rank(A) \le c$, equation (24) satisfies
\begin{equation} \ rank({x_i}A{x_j}^T) \le \min ({r_i},{c},{r_j}) \end{equation}
when $c$ is less than ${r_i}$ and ${r_j}$, equation (25) is written as
\begin{equation} \ rank({x_i}A{x_j}^T) \le {c} \end{equation}
Otherwise, equation (25) satisfies
\begin{equation} \ rank({x_i}A{x_j}^T) \le min({r_i,r_j}) <\ {c} \end{equation}

It can be shown that the solution to equation (6) is in the form of:
{\small
\begin{equation}
\begin{array}{l}
{x_1}A{x_2}^T{\omega _2} + {x_1}A{x_3}^T{\omega _3} +  \cdots + {x_1}A{x_P}^T{\omega _P} = {x_1}{x_1}^T{\omega _1}\\
{x_2}A{x_1}^T{\omega _1} + {x_2}A{x_3}^T{\omega _3} +  \cdots + {x_2}A{x_P}^T{\omega _P} = {x_2}{x_2}^T{\omega _2}\\
                \centerline \vdots \\
{x_P}A{x_1}^T{\omega _1} + {x_P}A{x_2}^T{\omega _2} +  \cdots + {x_P}A{x_{P - 1}}^T{\omega _{P - 1}} = {x_P}{x_P}^T{\omega _P}
\end{array}
\end{equation}}
Since $A$ is a diagonal matrix and the diagonal element is ${H_{{n_{i{1}}} \times {n_{i1}}}}$, equation (28) is rewritten as following form:
\begin{equation}
\begin{array}{l}
{x_1}A({x_2}^T{\omega _2} + {x_3}^T{\omega _3} +  \cdots  + {x_P}^T{\omega _P}) = {x_1}{x_1}^T{\omega _1}\\
{x_2}A({x_1}^T{\omega _1} + {x_3}^T{\omega _3} +  \cdots  + {x_P}^T{\omega _P}) = {x_2}{x_2}^T{\omega _2}\\
                \centerline \vdots \\
{x_P}A({x_1}^T{\omega _1} + {x_2}^T{\omega _2} +  \cdots  + {x_{P - 1}}^T{\omega _{P - 1}}) = {x_P}{x_P}^T{\omega _P}
\end{array}
\end{equation}
The rank of $\omega$ satisfies the following equation:
\begin{equation} \ rank(\omega) \le c \end{equation}
Since when the value of ${\rho _i}$ greater than zero, the corresponding projected vector contributes positively to the discriminative power in classification. Therefore, when $\omega ^T$ is in the form of equation (31), it will achieve the optimal recognition accuracy.
\begin{equation}
{\omega ^T} = {R^{d \times Q}}(d \le c)
\end{equation}
\begin{equation}
{X} = {R^{Q \times n}}
\end{equation}
Thus, the projected vector satisfies the following relation:
\begin{equation}
Y = {\omega ^T}X = {\omega ^T}\left[ \begin{array}{l}
{x_1}\\
{x_2}\\
 \vdots \\
{x_P}
\end{array} \right] = {R^{d \times n}}  (d \le c)
\end{equation}

Thus, expressions in (33) lead to the proof of (11).\\\indent

%



\ifCLASSOPTIONcompsoc
  \section*{Acknowledgments}
\else
  \section*{Acknowledgment}
\fi

This work is supported by the National Natural Science
Foundation of China (NSFC, No.61071211), the State Key
Program of NSFC (No. 61331201), the Key International
Collaboration Program of NSFC (No. 61210005) and the
Discovery Grant of Natural Science and Engineering Council
of Canada (No. 238813/2010).

\ifCLASSOPTIONcaptionsoff
  \newpage
\fi

\end{document}